\documentclass{llncs}

\usepackage[T1]{fontenc}
\usepackage[utf8]{inputenc}
\usepackage{graphicx}
\usepackage{amsmath,amssymb}
\usepackage{booktabs}
\usepackage{array}
\usepackage{url}

\graphicspath{{img/}}

\newcommand{\z}{\phantom{0}}
\newcolumntype{L}[1]{>{\raggedright\arraybackslash}p{#1}}

\begin{document}

\title{GRACE: Controlled Evidence Interventions for Diagnosing and Improving Knowledge Integration in GraphRAG}
\titlerunning{GRACE: Diagnosing and Improving Knowledge Integration}

\author{
Minh Pham \and
Quang Nguyen \and
Phong Kieu \and
Cong~Tran\orcidID{0000-0001-9467-4978}\textsuperscript{*}
}

\authorrunning{M. Pham et al.}

\institute{
Posts and Telecommunications Institute of Technology, Hanoi, Vietnam\\
\email{\{MinhPV.B22KH078, PhongKH.B22KH084\}@stu.ptit.edu.vn}\\
\email{quangnm@ptit.edu.vn}\\
\textsuperscript{*}Corresponding author:
\email{congtt@ptit.edu.vn}
}

\maketitle

\pagestyle{empty}
\thispagestyle{empty}

\begin{abstract}

GraphRAG systems answer knowledge-intensive questions by combining a language model's parametric knowledge with relational evidence retrieved from a graph. Aggregate accuracy, however, obscures whether failures arise from missing paths, distractors, or preference for memorised knowledge. Diagnosing these behaviours requires changing the evidence while holding the question fixed. We introduce GRACE, constructed by executing LC-QuAD~2.0 SPARQL queries over Wikidata to recover answers and supporting paths, extracting an intact evidence subgraph for each question, and deriving four controlled interventions from it: severing a supporting path, injecting type-compatible distractors, attaching high-degree hubs, and replacing the supported answer with a counterfactual entity. All five states share the same question and option positions, with targets derived from their evidence; retaining only validated five-state groups yields 4,028 questions and 20,140 question--graph pairs. These comparisons expose two limitations: question-only routing cannot distinguish different evidence states for the same question, and relevance guidance improves answer-supporting states but reduces accuracy when the support path is severed. The latter shows that individually useful facts do not necessarily provide sufficient collective support. This motivates QCFSE-SA, which jointly models question--fact relevance and graph-level sufficiency. Typed local attention follows connections between facts, global attention integrates the complete supplied graph, and question-conditioned readouts provide organisational guidance to a frozen generator. A separately calibrated graph-support gate controls abstention, addressing insufficiency that relevance guidance alone does not resolve. With identical factual context and a frozen generator, QCFSE-SA raises controlled-state accuracy from 57.17\% to 62.78\% (+5.61 points; question-clustered 95\% CI [+5.04,+6.17]) and broken-state accuracy from 55.41\% to 67.85\%, although over-abstention remains a challenge on semantic distractors.

\keywords{Retrieval-augmented generation \and Knowledge graphs \and Benchmark
\and Causal evaluation \and Knowledge conflict.}
\end{abstract}

\section{Introduction}

Retrieval-augmented generation (RAG) combines parametric knowledge with inference-time evidence \cite{lewis2020rag,gao2023ragsurvey}, while graph-based variants organise that evidence as entities, relations, or hierarchical summaries \cite{edge2024graphrag,guo2025lightrag,sarthi2024raptor}. This combination creates a basic attribution problem: a correct answer does not reveal whether the generator actually used the retrieved graph or simply relied on knowledge already encoded in its parameters. Consider the query \emph{“What currency is used by the country whose capital is Paris?”}. A language model may already associate France with the euro, while a retrieved graph may independently support the same answer through the path Paris $\rightarrow$ France $\rightarrow$ Euro. If the system outputs Euro, correctness alone cannot tell which source determined the prediction.

This ambiguity becomes consequential when the retrieved evidence is less straightforward. The graph may contain facts that are individually relevant but collectively insufficient to support an answer, plausible alternatives such as French franc connected through a different relation, or highly connected entities that are structurally salient without supporting the query. In other cases, the supplied graph may support an answer that conflicts with the model's parametric knowledge. The central question is therefore not only whether GraphRAG answers correctly, but whether it can distinguish supporting evidence from distraction, recognise when the available evidence is insufficient, and follow the supplied context when it conflicts with memory. A single end-to-end score cannot distinguish these failure modes. Text perturbation studies have exposed related robustness failures \cite{chen2024rgb,cuconasu2024noise,yoran2024robust}, but do not isolate the structural failure modes of graph evidence.

GRACE addresses this gap through controlled, matched evidence conditions. For each question, it fixes the question and answer interface while evaluating the system under five graph states that separate complete support, missing support, semantic distraction, topological distraction, and context--memory conflict. Holding the question fixed removes variation in question difficulty from these comparisons and also exposes a limitation of pre-retrieval, question-only routing: the same question can be paired with evidence that is sufficient, insufficient, distracting, or conflicting, none of which can be observed before the retrieved graph is inspected.

More importantly, these controlled comparisons reveal that relevance is not sufficiency. In our experiments, relevance-based evidence organisation improves every state in which the graph still supports an answer, yet reduces accuracy when the critical support path is severed: residual facts can remain relevant to the question without jointly forming a complete basis for answering. This observation motivates two complementary components. QCFSE organises the complete supplied graph conditional on the question and its fact connections, while QCFSE-SA additionally estimates whether the graph as a whole provides sufficient support to answer. The factual context and frozen generator remain identical in the paired comparison, isolating evidence integration from changes in retrieval volume or model scale.

\medskip\noindent\textbf{Research Questions.}

We ask which failure modes the controlled interventions expose (RQ1), whether question-only routing can distinguish evidence conditions that arise after retrieval (RQ2), and whether question-conditioned evidence organisation and explicit sufficiency control improve a frozen generator under fixed factual context (RQ3).

\medskip\noindent\textbf{Contributions.}
\begin{itemize}
\item GRACE: 4,028 questions and 20,140 matched observations across five graph
states, with a validated construction and evaluation protocol.
\item A five-state audit of four graph-retrieval frameworks and two
question-only routing policies.
\item QCFSE, which produces question-conditioned organisational guidance over the complete supplied graph while preserving its factual context, and QCFSE-SA, which adds calibrated graph-level sufficiency control.
\item The empirical finding that relevance and sufficiency are distinct requirements.
\end{itemize}

\section{Related Work}

\subsection{GraphRAG and KGQA}

Retrieval-augmented generation combined learned retrieval with generation and
multi-passage reading
\cite{karpukhin2020dpr,guu2020realm,lewis2020rag,izacard2021fid,borgeaud2022retro,izacard2023atlas,gao2023ragsurvey}.
Later systems organised evidence through corpus graphs, hierarchical summaries,
or associative retrieval
\cite{edge2024graphrag,guo2025lightrag,sarthi2024raptor,gutierrez2024hipporag},
while KG-oriented methods exposed structured interfaces, retrieved relation
paths, traversed graphs, or selected and scored question-specific subgraphs
\cite{jiang2023structgpt,luo2024rog,sun2024tog,he2024gretriever,mavromatis2025gnnrag,cheng2026faico,pan2024unifying,peng2024graphragsurvey,zhang2025graphragsurvey}.
KG-completion work likewise addressed missing structure
\cite{lin2024structkgc,liu2025sat}. Existing KGQA and knowledge-intensive
benchmarks covered semantic parsing, compositional questions, verification, and
shared-corpus retrieval
\cite{dubey2019lcquad,vrandecic2014wikidata,yih2016webqsp,talmor2018cwq,kim2023factkg,petroni2021kilt,yang2024crag,es2024ragas},
but ordinarily associated each question with only one evidence condition.

\subsection{Robustness and Knowledge Conflict}

Prior robustness studies separated noise rejection, integration, and
counterfactual behaviour, and showed that irrelevant content, its position, or
even a small adversarial injection could change a RAG answer
\cite{chen2024rgb,cuconasu2024noise,yoran2024robust,liu2024lostmiddle,shi2023distracted,zou2025poisonedrag}.
A complementary literature examined disagreement between parametric memory and
context, including entity substitution, explicit separation of contextual and
memorised answers, conflict taxonomies, and unreliable self-assessment of
knowledge boundaries
\cite{longpre2021entity,xie2024chameleon,neeman2023disentqa,xu2024conflictsurvey,ren2025boundary}.
These studies primarily perturbed passages or inferred conflict behaviour from
separate examples.

\subsection{Adaptive Retrieval and Controlled Evaluation}

Adaptive RAG methods triggered, skipped, or iterated retrieval using generation
confidence, reflection tokens, predicted complexity, self-reported knowledge,
reasoning steps, or entity popularity
\cite{jiang2023flare,asai2024selfrag,jeong2024adaptiverag,wang2023skr,trivedi2023ircot,mallen2023when};
these decisions inherited known difficulties in confidence calibration,
self-evaluation, and selective prediction
\cite{guo2017calibration,kadavath2022know,kamath2020selective}.
Controlled evaluation offered a complementary methodology: behavioural tests,
contrast sets, and counterfactually augmented data isolated targeted decision
factors rather than relying only on aggregate accuracy
\cite{ribeiro2020checklist,gardner2020contrast,kaushik2020counterfactual}.

\subsection{Research Gap and Positioning of GRACE}

The preceding lines of work leave three evaluation factors entangled. Standard
KGQA benchmarks vary the question together with its evidence, passage-level
robustness tests do not capture failures caused by graph structure, and adaptive
retrieval policies often decide from the question or model confidence before
observing the evidence that must support the answer. Consequently, they cannot
cleanly determine whether an error arises from retrieval routing, evidence
organisation, or failure to recognise that the available evidence is
insufficient.

GRACE addresses this gap by holding the question, answer options, and option
positions fixed while intervening directly on the supplied graph. For every
question, it evaluates an intact graph alongside states that sever the reasoning
path, add type-compatible or topologically salient distractors, or replace the
supported entity. These matched comparisons isolate missing support,
graph-specific distraction, and context--memory conflict from question
difficulty. They also allow a question-only gate to be evaluated against the
graph conditions realised after routing, revealing when an evidence-dependent
decision cannot be made from the question alone and motivating separate
mechanisms for evidence organisation and sufficiency recognition.

\section{The GRACE Benchmark}
\label{sec:benchmark}

\subsection{Matched Construction and Source Data}

GRACE treats the five states of one question as a matched group. For question
$q$, the natural-language text, the ordered option tuple $O_q$, and the option
letters remain fixed; only the supplied graph $S_{q,v}$ and its state-dependent
target $y_{q,v}$ change. This design reduces confounding by question difficulty
and makes within-question prediction changes attributable to the evidence
intervention.

\begin{figure}[t]
\centering
\includegraphics[width=\linewidth]{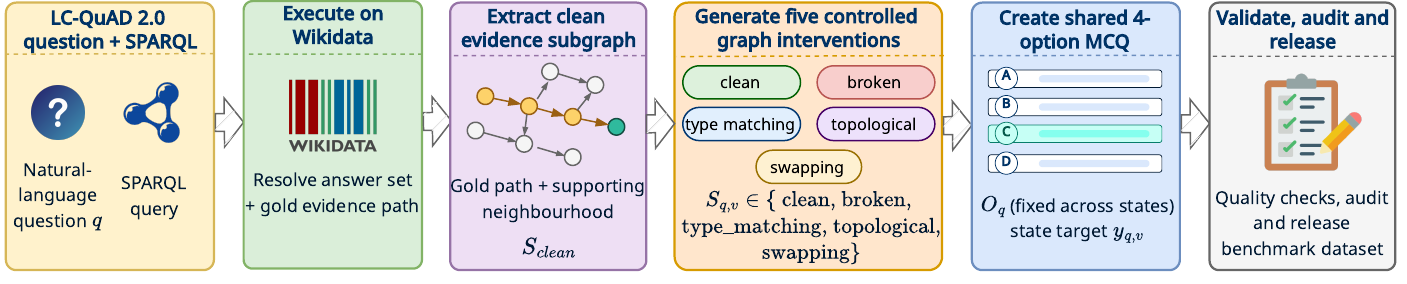}
\caption{GRACE construction pipeline. The question and option interface remain
fixed while the evidence graph is transformed into five matched states and
validated as one group.}
\label{fig:pipeline}
\end{figure}

Questions and executable SPARQL programs come from LC-QuAD~2.0
\cite{dubey2019lcquad}.\footnote{\url{http://lc-quad.sda.tech/}} We execute
\texttt{SELECT} queries against Wikidata \cite{vrandecic2014wikidata}, resolve
the answer and supporting path, map identifiers to English labels, and extract
a clean subgraph containing that path and its neighbourhood. As
Fig.~\ref{fig:pipeline} shows, every retained source item must yield all five
states, one fixed option tuple, and a fully validated mapping.

\subsection{Five Evidence States}

Table~\ref{tab:variants} defines five transformations of the same clean graph.
They form matched conditions rather than five unrelated datasets.


\begin{table}[!tp]
\centering
\caption{Controlled graph states and the behaviour each state isolates. All examples use the query ``What currency is used by the country whose capital is Paris?''}
\label{tab:variants}
\small
\setlength{\tabcolsep}{2pt}
\renewcommand{\arraystretch}{1.05}
\begin{tabular}{@{}lL{2.45cm}>{\centering\arraybackslash}m{4.15cm}L{3.05cm}@{}}
\toprule
State & Graph transformation & Example & Diagnostic question \\
\midrule
\emph{clean} & Keep the resolved evidence path and neighbourhood intact. & \begin{minipage}[c][2.75cm][s]{4.05cm}\centering\vspace*{\fill}\includegraphics[width=3.95cm,height=1.80cm,keepaspectratio]{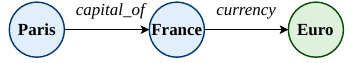}\vspace*{\fill}\par{\footnotesize\emph{Expected: Euro}}\end{minipage} & Can the system use complete, correct graph evidence? \\
\addlinespace[2pt]
\emph{broken} & Remove a critical gold-path triple (edge), severing the resolved support path.& \begin{minipage}[c][2.75cm][s]{4.05cm}\centering\vspace*{\fill}\includegraphics[width=3.95cm,height=1.80cm,keepaspectratio]{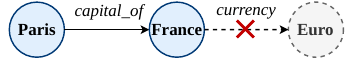}\vspace*{\fill}\par{\footnotesize\emph{Expected: Insufficient information}}\end{minipage} & Can it recognise that related residual facts do not complete a proof? \\
\addlinespace[2pt]
\emph{type matching} & Add entities compatible with the semantic type of the answer. & \begin{minipage}[c][2.90cm][s]{4.05cm}\centering\vspace*{\fill}\includegraphics[width=3.95cm,height=2.00cm,keepaspectratio]{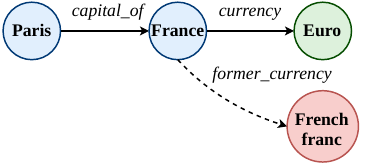}\vspace*{\fill}\par{\footnotesize\emph{Expected: Euro}}\end{minipage} & Can it resist plausible semantic alternatives? \\
\addlinespace[2pt]
\emph{topological} & Add high-degree entities and their connections. & \begin{minipage}[c][3.85cm][c]{4.05cm}\centering\includegraphics[width=4.05cm,height=2.55cm,keepaspectratio]{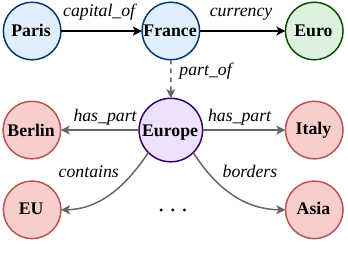}\par\vspace{3pt}{\footnotesize\emph{Expected: Euro}}\end{minipage} & Can it resist structural salience and hub bias? \\
\addlinespace[2pt]
\emph{swapping} & Substitute the supported answer entity with a distinct entity while preserving a coherent graph form. & \begin{minipage}[c][2.75cm][s]{4.05cm}\centering\vspace*{\fill}\includegraphics[width=3.95cm,height=1.80cm,keepaspectratio]{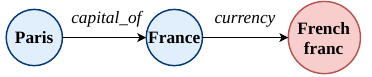}\vspace*{\fill}\par{\footnotesize\emph{Expected: French franc}}\end{minipage} & Does it update toward the supplied graph when context conflicts with parametric memory? \\
\bottomrule
\end{tabular}
\end{table}

The two diagnostically unusual cases are deliberate. \emph{broken} preserves
related residual facts but removes a necessary link, so it tests insufficiency
rather than generic noise rejection. \emph{swapping} remains internally
answerable and tests consistency with supplied evidence rather than recovery of
an external-world fact.

\subsection{Fixed Options and Controlled-State Targets}

Each four-option tuple contains the factual answer, the entity introduced by
\emph{swapping}, \emph{Insufficient information to answer}, and one additional
distractor. Distinct strings are shuffled once with a question-derived seed and
then reused across states. Let $y_q^{\mathrm{fact}}$,
$y_q^{\mathrm{ins}}$, and $y_q^{\mathrm{swap}}$ be their respective option
letters. The target is
\begin{equation}
y_{q,v}=\begin{cases}
y_q^{\mathrm{fact}}, & v\in\{\text{clean},\text{type matching},
\text{topological}\},\\
y_q^{\mathrm{ins}}, & v=\text{broken},\\
y_q^{\mathrm{swap}}, & v=\text{swapping}.
\end{cases}
\label{eq:state-targets}
\end{equation}
The clean state anchors factual QA; the controlled average additionally measures
abstention after evidence removal and response to an explicit counterfactual.
We therefore call it controlled-state accuracy rather than external-world
factual accuracy.

\subsection{Validation and Statistics}

A group is accepted only if the clean path is well formed, all transformations
succeed, the swapped entity is distinct and present, four options map
bijectively to A--D, and every answer and transformation remains traceable to
its source. Missing or unparsable system outputs count as errors. Of 6,046 test
candidates, 4,028 pass all checks, yielding 20,140 observations with complete
five-state coverage (Table~\ref{tab:stats}). The same rules create the separate
training resources.

\begin{table}[t]
\centering
\caption{GRACE test-set statistics. Percentages describe retained questions.}
\label{tab:stats}
\small
\setlength{\tabcolsep}{5.5pt}
\begin{tabular}{@{}lr@{\hspace{8mm}}lr@{}}
\toprule
Property & Value & Question category & Share \\
\midrule
Source candidates & 6,046 & timeless fact & 51.3\% \\
Retained questions & 4,028 & past event, no timestamp & 33.0\% \\
States per question & 5 & past event with timestamp & \z8.3\% \\
Question--graph pairs & 20,140 & present status & \z4.7\% \\
Matched-state coverage & 100\% & fictional or mythological & \z2.8\% \\
SPARQL form & \texttt{SELECT} & validated structural errors & 0 \\
\bottomrule
\end{tabular}
\end{table}

The intervention is controlled at the graph-input level, not at every internal
stage of a retrieval framework; indexing, retrieval, and context construction
can still introduce system-specific failures. The paired QCFSE experiment adds
a stricter byte-identical factual-context control in Section~\ref{sec:experiment}.

\section{Proposed Method}
\label{sec:method}

\subsection{Overview and Decision Rule}

A retrieved graph creates two distinct decisions. \emph{Evidence organisation}
asks which facts and connections matter for the question; \emph{evidence
sufficiency} asks whether those facts form a complete basis for answering. A
remaining fact can be relevant after the decisive path has been severed, so
relevance alone cannot justify an answer. QCFSE addresses the first decision by
encoding the complete question--graph pair, and QCFSE-SA adds an explicit
sufficiency controller.

Let $q$ be a question, $O_q$ its ordered options, and
$S_v=(e_1,\ldots,e_n)$ the complete graph supplied in evidence state $v$, where
$e_i=(s_i,r_i,o_i)$ is a KG triple. QCFSE produces organisational guidance
$g_\theta(q,S_v)$ and a graph-support probability $c_\theta(q,S_v)\in[0,1]$.
With frozen generator $G$, calibration-selected threshold $\tau$, and
insufficient-information option $y_{\mathrm{ins}}$, the complete decision is
\begin{equation}
\hat y_{q,v}=
\begin{cases}
 y_{\mathrm{ins}}, & c_\theta(q,S_v)<\tau,\\
 G\!\left(q,O_q,S_v,g_\theta(q,S_v)\right),
 & c_\theta(q,S_v)\geq\tau.
\end{cases}
\label{eq:qcfse-sa}
\end{equation}
QCFSE always takes the second branch; QCFSE-SA may abstain. Both inspect every
fact, and $G$ receives the same ordered factual context as the Full-KG control.

\begin{figure}[t]
\centering
\includegraphics[width=\linewidth]{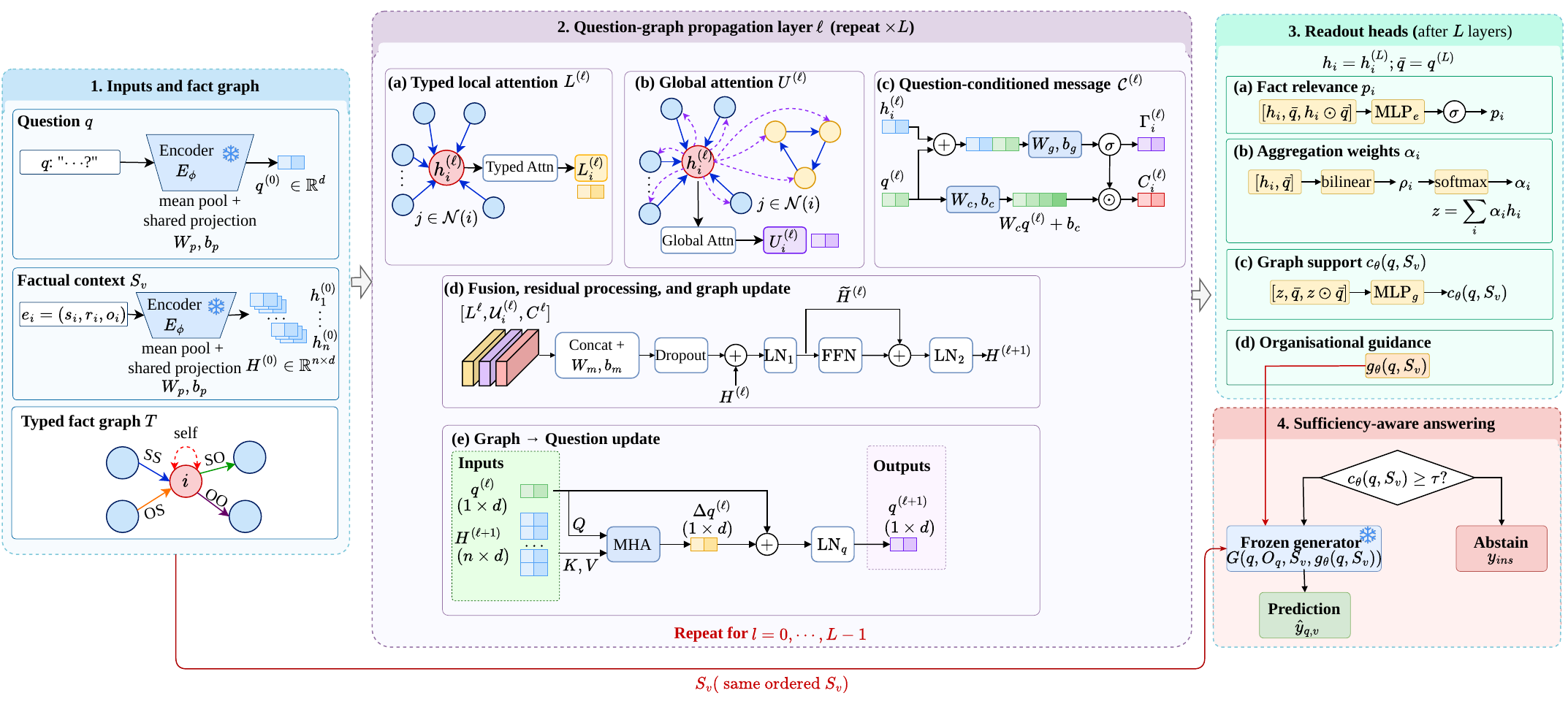}
\caption{QCFSE-SA architecture. Typed local attention follows fact-graph edges,
global attention compares all facts, and question-conditioned messages couple
the two representations. Separate readouts estimate fact relevance, aggregation
weights, and graph support before sufficiency-aware answering.}
\label{fig:qcfse-architecture}
\end{figure}

Figure~\ref{fig:qcfse-architecture} summarises inference and the separate
training/calibration stages.

\subsection{Question-Conditioned Full-Subgraph Propagation}

Each triple becomes a fact node. Two nodes are connected when their subject or
object endpoints coincide. The directed edge type
$T_{ij}\in\{SS,SO,OS,OO,\mathrm{self},\varnothing\}$ records which endpoints
match; $\mathcal N(i)=\{j:T_{ij}\neq\varnothing\}$ is the local neighbourhood.
A frozen semantic encoder $E_\phi$ embeds the question and verbalised triples,
and a learned projection maps them to initial states
$q^{(0)}\in\mathbb R^d$ and
$H^{(0)}=[h_1^{(0)},\ldots,h_n^{(0)}]\in\mathbb R^{n\times d}$.

At propagation layer $\ell$, typed local attention follows only graph-adjacent
facts. For head $a$, with $d_h=d/H$, it uses
\begin{align}
s_{ij}^{a}&=\frac{\langle h_i^{(\ell)}W_Q^a,
 h_j^{(\ell)}W_K^a\rangle}{\sqrt{d_h}}+b_{T_{ij}}^a,\\
a_{ij}^{a}&=\operatorname{softmax}_{j\in\mathcal N(i)}(s_{ij}^{a}),\\
L_i^{(\ell)}&=W_O\operatorname*{Concat}_{a=1}^{H}
 \left(\sum_{j\in\mathcal N(i)}a_{ij}^{a}h_j^{(\ell)}W_V^a\right).
\label{eq:typed-local-compact}
\end{align}
The learned scalar $b_{T_{ij}}^a$ lets structural role affect attention
independently of semantic similarity. In contrast, the global branch
$U^{(\ell)}=\operatorname{MHA}_{\mathrm{global}}(H^{(\ell)})$ uses standard
self-attention with a separate set of projections, normalises each head over
\emph{all} $j\in\{1,\ldots,n\}$, and uses neither the adjacency mask nor an
edge-type bias. Thus $L^{(\ell)}$ preserves endpoint-sharing paths, whereas
$U^{(\ell)}$ compares locally disconnected facts.

A feature-wise gate injects the current question into each fact,
\begin{align}
\Gamma_i^{(\ell)}&=\sigma\!\left(W_g
 \operatorname{Concat}(h_i^{(\ell)},q^{(\ell)})+b_g\right),\\
C_i^{(\ell)}&=\Gamma_i^{(\ell)}\odot(W_cq^{(\ell)}+b_c).
\label{eq:question-message-compact}
\end{align}
The local, global, and question-conditioned messages are fused by feature
concatenation:
\begin{align}
R^{(\ell)}&=W_m\operatorname{Concat}
 \left(L^{(\ell)},U^{(\ell)},C^{(\ell)}\right)+b_m,\\
H^{(\ell+1)}&=\operatorname{TransformerBlock}
 \left(H^{(\ell)},R^{(\ell)}\right),\\
q^{(\ell+1)}&=\operatorname{LN}\!\left(q^{(\ell)}+
 \operatorname{MHA}(q^{(\ell)},H^{(\ell+1)},H^{(\ell+1)})\right).
\label{eq:bidirectional-update}
\end{align}
The transformer block comprises the standard residual, layer-normalisation, and
feed-forward operations. The last line writes the updated graph back into the
question, yielding a bidirectional $q\rightarrow S_v\rightarrow q$ interaction
at every layer.

\subsection{Relevance, Aggregation, and Sufficiency Readouts}

After the final layer, write $h_i=h_i^{(L)}$ and $\bar q=q^{(L)}$. A
question-conditioned softmax produces aggregation weights $\alpha_i$ and graph
summary $z$:
\begin{equation}
\rho_i=\frac{\langle W_fh_i,W_q\bar q\rangle}{\sqrt d},\qquad
\alpha_i=\frac{e^{\rho_i}}{\sum_j e^{\rho_j}},\qquad
z=\sum_i\alpha_i h_i.
\label{eq:compact-pooling}
\end{equation}
Separate heads then predict independent fact relevance and whole-graph support,
\begin{align}
p_i&=\sigma\!\left(\operatorname{MLP}_e(
 \operatorname{Concat}(h_i,\bar q,h_i\odot\bar q))\right),\\
c_\theta(q,S_v)&=\sigma\!\left(\operatorname{MLP}_g(
 \operatorname{Concat}(z,\bar q,z\odot\bar q))\right).
\label{eq:compact-heads}
\end{align}
The distinction is essential: $\alpha$ distributes unit mass for aggregation,
each $p_i$ independently marks a potentially critical fact, and $c_\theta$
judges the graph as a whole. QCFSE serialises these scores and observed fact
connections as soft metadata beside the unchanged graph. The metadata contains
no option letters, candidate answers, labels, or generator outputs.

\subsection{Structural Training Objective}

Training uses only questions and paired clean/broken graphs. For question $q$,
let $C_q$ and $B_q$ be those fact sets, let
$y_i^e=\mathbf 1[e_i\in C_q\setminus B_q]$ mark a fact removed by path
severing, and let $q'$ be a mismatched question from the minibatch. The fact
head logit is $\xi_i$; the graph-support logit is $\gamma(q,S)$, with
$c_\theta(q,S)=\sigma(\gamma(q,S))$. The complete objective is
\begin{align}
\mathcal L_e&=-\frac1{N_{\mathrm{lab}}}\sum_i
 \left[w_+y_i^e\log\sigma(\xi_i)+(1-y_i^e)
 \log(1-\sigma(\xi_i))\right],\\
\mathcal L_{cb}&=\frac1B\sum_q\operatorname{softplus}
 \left(m-\gamma(q,C_q)+\gamma(q,B_q)\right),\\
\mathcal L_{qm}&=\frac1B\sum_q\operatorname{softplus}
 \left(m-\gamma(q,C_q)+\gamma(q',C_q)\right),\\
\mathcal L&=\lambda_e\mathcal L_e+\lambda_{cb}\mathcal L_{cb}
 +\lambda_{qm}\mathcal L_{qm}.
\label{eq:compact-objective}
\end{align}
The three losses are complementary. $\mathcal L_e$ identifies locally critical
facts and uses positive weight $w_+$ for class imbalance. $\mathcal L_{cb}$
encourages $\gamma(q,C_q)\geq\gamma(q,B_q)+m$, teaching whether the controlled
path remains sufficient. $\mathcal L_{qm}$ encourages
$\gamma(q,C_q)\geq\gamma(q',C_q)+m$, preventing the support head from treating
generic graph completeness as question-specific support. Here $m>0$ is the
required ranking margin and the $\lambda$ values weight the objectives. The
encoder receives no answer options, answer keys, or generator responses.

\subsection{Sufficiency Calibration and Context Preservation}
\label{sec:calibration}

After training and checkpoint selection, $\tau$ is selected on a disjoint
300-question calibration set by maximising controlled-state accuracy over
observed support scores and is frozen before testing. Calibration labels
therefore determine only the decision threshold and do not train the evidence
representation. The paired comparison also preserves factual context: both arms
pass the identical serialised $S_v$ to $G$, only $g_\theta$ is appended, and
ordered-context hashes are checked for every pair.

\section{Experimental Evaluation}
\label{sec:experiment}

\subsection{Metrics and Protocol}
\label{sec:setup}



We report per-state exact-match accuracy and their equal-weight mean
(controlled-state accuracy); failed or unparsable outputs count as incorrect.
All 95\% confidence intervals use 10,000 paired question-clustered bootstrap
resamples, keeping the five states of each question in one cluster. QCFSE uses 4,250 questions for training, 750 for checkpoint selection, and a
disjoint 300-question calibration set for sufficiency-threshold selection. The common generator is Qwen3.5-35B-A3B
(4-bit AWQ, temperature 0, seed 42), and the frozen encoder is
\texttt{cross-encoder/ms-marco-MiniLM-L6-v2}. All paired Full-KG/QCFSE runs fix
the question, options, ordered graph, generator, prompt, and decoding settings;
factual-context hashes match for all 20,140 test instances.

\subsection{Retrieval-Framework Baselines}
\label{sec:baselines}

GraphRAG \cite{edge2024graphrag}, LightRAG \cite{guo2025lightrag}, RAPTOR
\cite{sarthi2024raptor}, and a FAICO-style retriever \cite{cheng2026faico} start
from the supplied state graph and feed their selected context to the common MCQ
generator and parser. They retain framework-specific community, local-graph,
hierarchical, or connected-subgraph indexing, but cannot access a different
corpus. Table~\ref{tab:baselines} reports per-state accuracy and completion for the
four retrieval-framework baselines.

\begin{table}[t]
\centering
\caption{Controlled-state accuracy (\%) of retrieval-framework artifacts.
Failures count as wrong; completion is the percentage of parsed state outputs.}
\label{tab:baselines}
\setlength{\tabcolsep}{2.5pt}
\begin{tabular}{@{}lrrrrrrrr@{}}
\toprule
System & $N$ & Comp. & Clean & Broken & Type & Topo. & Swap & Avg. \\
\midrule
RAPTOR & 4,028 & 99.68 & 62.16 & 52.58 & 56.28 & 55.88 & 40.47 & 53.48 \\
FAICO-style & 4,028 & 100.00 & 31.33 & 69.86 & 31.63 & 31.26 & 19.71 & 36.76 \\
GraphRAG & 4,028 & 29.98 & 13.95 & 16.26 & 12.41 & 15.39 & 10.13 & 13.63 \\
LightRAG snapshot & 2,946 & 90.29 & 49.29 & 44.30 & 45.08 & 48.37 & 36.35 & 44.68 \\
\bottomrule
\end{tabular}
\end{table}

RAPTOR is strongest among completed general-purpose baselines but degrades most
under swapping. FAICO-style retrieval is conservative on broken graphs yet weak
on intact and conflicting evidence, showing that connectivity-based abstention
is not accurate evidence use. GraphRAG and LightRAG are primarily pipeline
diagnostics because of low and incomplete completion, respectively, and do not
support a universal framework ranking.

\subsection{Question-Only Answerability Routing}
\label{sec:routingresults}

Two DeBERTa-v3-base encoders~\cite{he2023debertav3} predict no-context
answerability from the question, using either open-ended answer matching or an
option-aligned probe. Table~\ref{tab:routing} compares them with always
retrieving.

\begin{table}[t]
\centering
\caption{Question-only routing over 4,028 questions. Retr. is the percentage
sent to the RAG branch.}
\label{tab:routing}
\small
\setlength{\tabcolsep}{3.5pt}
\begin{tabular}{@{}lrrrrrrr@{}}
\toprule
Method & Clean & Broken & Type & Topo. & Swap & Avg. & Retr. \\
\midrule
Always retrieve & 60.90 & 54.57 & 59.29 & 59.93 & 42.38 & 55.41 & 100.00 \\
Open-ended gate & \textbf{62.31} & \textbf{55.73} & \textbf{60.68} & \textbf{61.47} & \textbf{43.84} & \textbf{56.81} & 88.41 \\
Option-aligned gate & 59.53 & 55.56 & 58.49 & 59.16 & 32.72 & 53.09 & 31.58 \\
\bottomrule
\end{tabular}
\end{table}

The open-ended gate gains 1.40 points (95\% CI $[+0.94,+1.89]$) while saving
11.59\% of calls. Aggressive option-aligned routing saves 68.42\% but loses 2.32
points, including 9.66 on \emph{swapping}. A question-only gate applies the
same route to every evidence state and can therefore suppress context precisely
when the graph must override a strong parametric answer.

\subsection{Evidence Organisation and Sufficiency Control}
\label{sec:qcfseresults}

\begin{figure}[htbp]
\centering
\includegraphics[width=.88\linewidth]{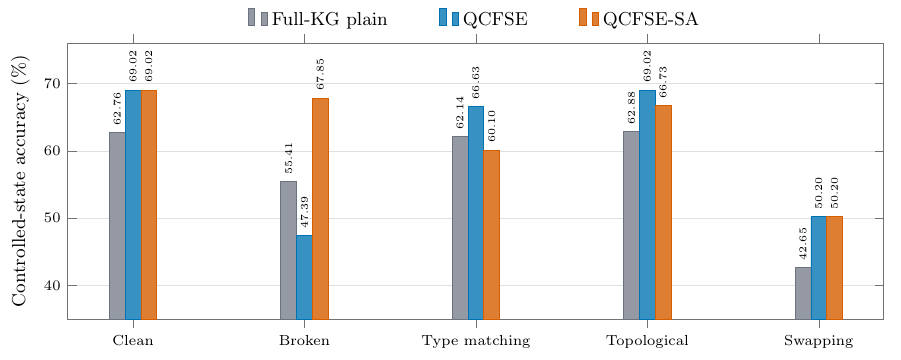}
\caption{Paired controlled-state accuracy. Evidence organisation helps on
supported graphs, while explicit sufficiency control repairs the severed-evidence
failure.}
\label{fig:teaser}
\end{figure}

Table~\ref{tab:qcfse} and Figure~\ref{fig:teaser} compare evidence organisation
alone with the addition of sufficiency control across the five graph states.

\begin{table}[t]
\centering
\caption{Paired full-test controlled-state accuracy (\%).}
\label{tab:qcfse}
\small
\setlength{\tabcolsep}{4.2pt}
\begin{tabular}{@{}lrrrrrr@{}}
\toprule
Method & Clean & Broken & Type & Topo. & Swap & Avg. \\
\midrule
Full-KG plain & 62.76 & 55.41 & 62.14 & 62.88 & 42.65 & 57.17 \\
QCFSE & \textbf{69.02} & 47.39 & \textbf{66.63} & \textbf{69.02} & \textbf{50.20} & 60.45 \\
QCFSE-SA & \textbf{69.02} & \textbf{67.85} & 60.10 & 66.73 & \textbf{50.20} & \textbf{62.78} \\
\bottomrule
\end{tabular}
\end{table}

\noindent\textbf{Relevance Is Not Sufficiency.}
QCFSE improves every supported-evidence state, with its largest gain on
\emph{swapping} (+7.55), but falls from 55.41\% to 47.39\% on
\emph{broken}. This does not contradict perfect clean-over-broken validation
ranking: ranking orders two graphs but does not make a broken score trigger
abstention. After severing, residual facts may remain relevant although they no
longer complete a proof; relevance guidance amplifies these clues while the
generator is still required to answer.

\noindent\textbf{Sufficiency-Aware Result.}
QCFSE-SA raises overall accuracy by 5.61 points over Full-KG plain, with
question-clustered 95\% CI $[+5.04,+6.17]$, and reaches 67.85\% on
\emph{broken} (+12.44). This repairs the objective mismatch while retaining
QCFSE's gains on clean and swapping graphs.

\noindent\textbf{Remaining Calibration Error.}
The gate also abstains on some answerable distractor graphs, causing
accuracy in the \emph{type matching} state to fall 2.04 points below plain. GRACE therefore
localises the remaining problem: distinguishing an incomplete proof from a
complete but difficult graph.

\subsection{Threats to Validity}

About 99\% of \emph{broken} serialisations contain an \emph{Unknown
entity} cue, which the method may exploit despite not seeing the state name.
GRACE is derived from one KGQA source (LC-QuAD 2.0 over Wikidata),
one query form (SELECT), and one paired generator;
incomplete LightRAG and GraphRAG runs preclude a final framework ranking.
Swapping measures consistency with supplied evidence, not external-world truth.
These caveats do not affect the byte-identical-context QCFSE comparison.

\section{Conclusion}

GRACE uses matched graph interventions to expose evidence-integration failures
that clean accuracy and question-only routing conceal. QCFSE shows that
question-conditioned organisation improves how a frozen generator uses a fixed
complete graph, while its broken-state regression establishes that relevance is
not sufficiency. QCFSE-SA makes sufficiency explicit, improving the aggregate
result while revealing over-abstention on difficult but answerable graphs as the
next calibration problem.

\bibliographystyle{splncs04}
\bibliography{references}

\clearpage

\end{document}